\documentclass[conference]{IEEEtran}
\IEEEoverridecommandlockouts
\usepackage{cite}
\usepackage{amsmath,amssymb,amsfonts}
\usepackage{graphicx}
\usepackage{textcomp}
\usepackage{xcolor}
\usepackage{url}
\usepackage{graphicx}
\usepackage{booktabs}
\usepackage{amsmath}
\usepackage{amssymb}
\usepackage{latexsym}
\usepackage{graphicx}
\usepackage{booktabs}
\usepackage{makecell}
\usepackage{longtable}
\usepackage{makecell}
\usepackage{hyperref}
\usepackage{framed,multirow}

\usepackage{amssymb}
\usepackage{latexsym}
\usepackage{graphicx}
\usepackage{booktabs}
\usepackage{makecell}
\usepackage{longtable}
\usepackage{makecell}
\usepackage{hyperref}
\usepackage{url}
\usepackage{xcolor}

\usepackage{hyperref}

\usepackage{graphicx}  
\usepackage{float}  
\usepackage{subfigure} 
\usepackage[linesnumbered,ruled,vlined]{algorithm2e}
\usepackage{algpseudocode}
\usepackage{amsmath}

\def\BibTeX{{\rm B\kern-.05em{\sc i\kern-.025em b}\kern-.08em
    T\kern-.1667em\lower.7ex\hbox{E}\kern-.125emX}}
\begin{document}
\setlength{\textfloatsep}{5pt} 

\title{\textit{GL-ICNN}: An End-To-End Interpretable Convolutional Neural Network for the Diagnosis and Prediction of Alzheimer's Disease\\
}
\author{\IEEEauthorblockN{Wenjie \snm{Kang}}
\IEEEauthorblockA{\textit{Biomedical Imaging Group Rotterdam\\Erasmus MC, NL\\w.kang@erasmusmc.nl}} \\
\and
\IEEEauthorblockN{Lize \snm{Jiskoot}}
\IEEEauthorblockA{\textit{Erasmus MC, NL} \\
}
\and
\IEEEauthorblockN{Peter De\snm{Deyn}}
\IEEEauthorblockA{\textit{University Medical Center Groningen, NL} \\
}
\and
\IEEEauthorblockN{Geert \snm{Biessels}}
\IEEEauthorblockA{\textit{University Medical Center Utrecht, NL} \\
}
\and
\IEEEauthorblockN{Huiberdina \snm{Koek}}
\IEEEauthorblockA{\textit{University Medical Center Utrecht, NL} \\
}
\and
\IEEEauthorblockN{Jurgen \snm{Claassen}}
\IEEEauthorblockA{\textit{Radboud University Medical Center, NL} \\
}
\and
\IEEEauthorblockN{Huub \snm{Middelkoop}}
\IEEEauthorblockA{\textit{Leiden University Medical Center, NL} \\
}
\and
\IEEEauthorblockN{Wiesje \snm{Flier}}
\IEEEauthorblockA{\textit{Amsterdam University Medical Center, NL} \\
}
\and
\IEEEauthorblockN{Willemijn \snm{Jansen}}
\IEEEauthorblockA{\textit{Maastricht University Medical Center, NL} \\
}
\and
\IEEEauthorblockN{Stefan \snm{Klein}}
\IEEEauthorblockA{\textit{Biomedical Imaging Group Rotterdam\\Erasmus MC, NL} \\
}
\and
\IEEEauthorblockN{Esther \snm{Bron}}
\IEEEauthorblockA{\textit{Biomedical Imaging Group Rotterdam\\Erasmus MC, NL} \\
}

}

\author{
\IEEEauthorblockN{
    Wenjie Kang$^{1,*}$, 
    Lize Jiskoot$^2$, 
    Peter De Deyn$^3$, 
    Geert Biessels$^4$, 
    Huiberdina Koek$^4$, \\
    Jurgen Claassen$^5$, 
    Huub Middelkoop$^6$, 
    Wiesje Flier$^7$, 
    Willemijn J. Jansen$^8$, 
    Stefan Klein$^1$, 
    Esther Bron$^1$
}
\IEEEauthorblockA{
    $^1$\textit{Biomedical Imaging Group Rotterdam, Erasmus MC, NL}; 
    $^2$\textit{Erasmus MC, NL}; 
    $^3$\textit{University Medical Center Groningen, NL}; \\
    $^4$\textit{University Medical Center Utrecht, NL}; 
    $^5$\textit{Radboud University Medical Center, NL}; \\
    $^6$\textit{Leiden University Medical Center, NL}; 
    $^7$\textit{Amsterdam University Medical Center, NL}; 
    $^8$\textit{Maastricht University Medical Center, NL}\\
    *w.kang@erasmusmc.nl
}
}

\maketitle

\begin{abstract}
Deep learning methods based on Convolutional Neural Networks (CNNs) have shown large potential to improve early and accurate diagnosis of Alzheimer’s disease (AD) dementia based on imaging data. However, these methods have yet to be widely adopted in clinical practice, possibly due to the limited interpretability of deep learning models. The Explainable Boosting Machine (EBM) is a glass-box model but cannot learn features directly from input imaging data. In this study, we propose a novel interpretable model that combines CNNs and EBMs for the diagnosis and prediction of AD. We develop an innovative training strategy that alternatingly trains the CNN component as a feature extractor and the EBM component as the output block to form an end-to-end model. The model takes imaging data as input and provides both predictions and interpretable feature importance measures. 
We validated the proposed model on the Alzheimer’s Disease Neuroimaging Initiative (ADNI) dataset, and the Health-RI Parelsnoer Neurodegenerative Diseases Biobank (PND) as an external testing set. The proposed model achieved an area-under-the-curve (AUC) of 0.956 for AD and control classification, and 0.694 for the prediction of conversion of mild cognitive impairment (MCI) to AD on the ADNI cohort. The proposed model is a glass-box model that achieves a comparable performance with other state-of-the-art black-box models. Our code is available at: \url{https://anonymous.4open.science/r/GL-ICNN}.

\end{abstract}


\section{Introduction}

Magnetic Resonance Imaging (MRI) has the potential to aid clinicians in differentiating Alzheimer’s disease (AD) dementia from other causes of mild cognitive impairment (MCI), and to help predict those at highest risk of progression to dementia \cite{ayers2019brain}. Convolutional Neural Networks (CNNs) have shown potential to improve diagnostic and prognostic yield from MRIs \cite{wen2020convolutional}. However, despite the promising performance of machine learning models, they have yet to be widely adopted in clinical practice \cite{martin2023interpretable}. A key reason is that those high-performance machine learning methods using imaging data are considered as black-box models, which are difficult to interpret \cite{leming2023challenges}. While the glass-box models, such as logistic regression are relatively easy for humans to interpret. \par

Explainable artificial intelligence (XAI) methods have been used to explain the outputs of CNNs \cite{arrieta2020explainable}. However, those post-hoc explanation methods have low fidelity \cite{kindermans2019reliability}. In our previous work, we integrated CNNs with Explainable Boosting Machine (EBM) \cite{lou2012intelligible} to build a framework that is both transparent and capable of leveraging high-dimensional brain images ({\textit{Glo\&Loc-EBM}}) \cite{kang2023interpretable}. EBM is an interpretable Generalize Additive Model (GAM) that provides feature importance estimates. However, the proposed framework is not an end-to-end model, which means the feature selection, feature extraction and prediction are in different steps. The complexity of the framework makes it hard to implement because of the high time consumption and task specific training strategy.\par

In this study, we propose an end-to-end interpretable model integrating CNN and EBM. Because it is challenging to train the model like a common CNN, we developed a novel training strategy that trains the model end-to-end by optimizing CNN and EBM components alternatingly in each epoch. The model considers features independently at both the whole-image (global) level and the brain region (local) level. The proposed model is called the Global and Local Interpretable Convolutional Neural Network (\textit{GL-ICNN}) in the rest of the paper. We validated \textit{GL-ICNN} on two cohorts, considering AD and control (CN) classification task and prediction of the conversion of MCI to AD task. The interpretable outputs provided by \textit{GL-ICNN} include individual-level and group-level feature importance, which helps people to understand how the model makes decisions and which brain regions play a role in AD diagnosis.\par

\section{Methods}
\subsection{\textit{GL-ICNN} architecture}

In our previous work \cite{kang2023interpretable}, we combined the advantages of CNNs in extracting high-dimensional features with the inherent interpretability of EBM. We propose here a more efficient and elegant end-to-end formulation than the {\textit{Glo\&Loc-EBM}}, which in particular eliminates the compute-intensive feature selection and extraction steps. The novel \textit{GL-ICNN} features an architecture that integrates multiple CNN backbones along with an EBM block. The CNN component extracts multi-scale features, while the EBM serves as the output block of the \textit{GL-ICNN}. Additionally, we introduce an innovative training strategy where the CNN and EBM components are trained alternatingly in each epoch.

The entire brain image is divided into non-overlapping patches that collectively cover the whole image. A CNN model is trained to take both the whole image (Global) and individual brain regions (Local) as input, with several fully connected layers serving as the output block. This model, referred to as \textit{GL-CNN}, is included in the comparison study as a black-box model. The CNN architecture consists of a global CNN backbone, which is a DenseNet \cite{huang2017densely}, and local CNN backbones which are based on VGG \cite{simonyan2014very}, this architecture was adapted from our previous study \cite{kang2023interpretable}.\par

\begin{figure}[h]
\centering
\includegraphics[width=0.45\textwidth]{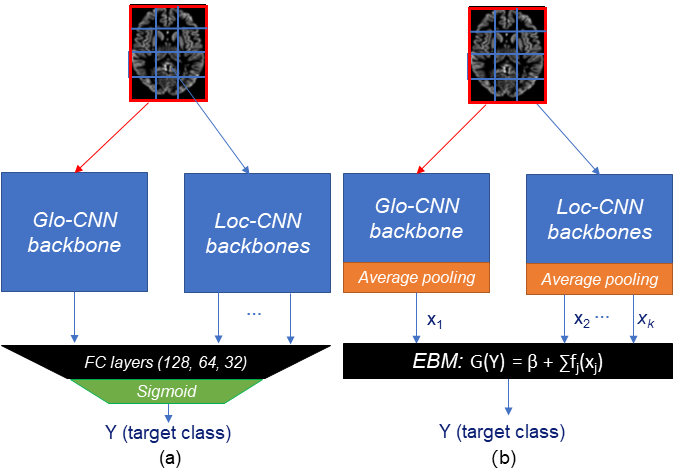}
\caption{The overview of (a) \textit{GL-CNN} (b) \textit{GL-ICNN}.} 
\label{overview}
\end{figure}

We initially train the \textit{GL-CNN} for several warm-up epochs, then we replace the fully connected layers of the \textit{GL-CNN} with an EBM and to construct the \textit{GL-ICNN}. We connect the CNN layers with the EBM by adding an average pooling layer at the end of the $j$th CNN, and defining its output as feature $x_j$ in the EBM. This complete \textit{GL-ICNN} model is subsequently trained in an end-to-end fashion as described in the subsection\,\ref{train}. The overview of the \textit{GL-ICNN} architecture is shown in Fig.\,\ref{overview}(b), where the outputs of CNN component in the \textit{GL-ICNN} ($Out_{CNN}$) are \{${x_1,...,x_N}$\}\label{test}. Given a dataset $D$ = \{$(X_i$, $y_i)$\}$_1^N$, where for any subject $i\in[1, N]$, {$X_i$} = $(x_{1}$, ..., $x_{k})$ is a feature vector with $k$ features, and $y_i$ is the label, EBM is of the form:
\begin{equation}\label{eqn-1}
g(y)=\beta + \sum f_j\left(x_j\right),
\end{equation}
\noindent
where $g(.)$ is the link function that adapts to classification setting, $\beta$ is the intercept, {$x_j$} is the $j$th feature, $y$ is the target class, and shape functions $f_j$ are gradient-boosted ensembles of bagged decision trees. Each shape function operates on a single variable, and shape functions are combined through summation, preventing interactions effects from being learned. Because of this ability of the EBM to analyze features independently, it is straightforward to interpret the contribution of each feature ($f()$ in equation \ref{eqn-1}) to the prediction. Because of the ability of EBM to analyze features independently, it is also straightforward to interpret the importance of each feature to the prediction \cite{caruana2015intelligible}. Hence, the individual-level feature importance vector for subject $i$ is defined as the collection of its shape function outputs:

\begin{equation}\label{eqn-2}
I_{indiv}(X_i) = \left(f_1(x_1), \ldots, f_k(x_k) \right)  
\end{equation}
\noindent
and group-level feature importance vector for a dataset $D$ is defined as the average of the absolute individual-level feature importance measures:
\begin{equation}\label{eqn-3}
I_{group}(D) = \frac{1}{N} \sum_{i=1}^{N} |I(X_i)|  
\end{equation}
\noindent
where $|\cdot|$ denotes the element-wise absolute operator.
\par

\subsection{End-to-end \textit{GL-ICNN} training}\label{train}
The decision trees in EBM are grown by repeatedly cycling through features with a small learning rate, which forces the model to sequentially consider each feature as an explanation of the current residual rather than greedily selecting the best feature (Algorithm\,2 in \cite{nori2021accuracy}). This particular training procedure of EBM makes it not straightforward to simply combine CNN and EBM in a neural network and rely on the usual backpropagation with stochastic gradient descent optimizer for end-to-end training. To address this, we propose a block coordinate descent approach, alternating between optimization of CNN and EBM weights. The training procedure of the \textit{GL-ICNN} is shown in Algorithm\,\ref{alg:AOA}. The hyperparameters, including ${N_{max}}$ and ${N_{tolerate}}$, are optimized using the validation set.

\begin{algorithm}[h]\label{alg:AOA}
\caption{Training of \textit{\textit{GL-ICNN}}}
\KwIn {\textit{GL-ICNN} with CNN weights resulting from \textit{GL-CNN} pre-training, hyperparameter ${N_{max}}$, hyperparameter ${N_{tolerate}}$}
\KwData {training set $D_{train}=\{X_{1...N_{train}}, y_{1...N_{train}}\}$, validation set $D_{valid}=\{X_{1...N_{valid}}, y_{1...N_{valid}}\}$}
\For{$epoch\,i\leftarrow 1$ \KwTo $N_{max}$}{
$\textit{\textit{GL-ICNN}}_i\leftarrow$ train the EBM component of the \textit{\textit{GL-ICNN}} using $Out_{CNN}$ while keeping the CNN weights fixed\par
$\textit{\textit{GL-ICNN}}_i\leftarrow$ train the CNN component of \textit{\textit{GL-ICNN}} while keeping EBM weights fixed\par
$loss_i\leftarrow$weighted cross entropy on $D_{valid}$\par
\lIf{$loss_i<loss_{i-1}$}{
\textit{\textit{GL-ICNN}}$_{best}\leftarrow$ save\,\textit{\textit{GL-ICNN}}$_{i}$, t=0}
\lElse{t=t+1}
\lIf{$t>N_{tolerate}$}{\textbf{break}}}

\textbf{return} \textit{GL-ICNN}$_{\textit{best}}$
\end{algorithm}

\section{Experiments} 
\subsection{Study materials}
We included participants with T1-weighted (T1w) MRI scans available at the baseline timepoint from the Alzheimer’s Disease Neuroimaging Initiative (ADNI) and the Health-RI Parelsnoer Neurodegenerative Diseases Biobank (PND) cohorts \cite{mannien2017parelsnoer}. The ADNI cohort consists of 335 AD patients, 520 control participants (CN), 231 MCI patients who converted to AD within 3 years (MCIc), and 629 MCI patients who did not convert (MCInc) within 3 years. The PND cohorts includes 198 AD patients, 138 participants with Subjective Cognitive Decline (SCD), 48 MCIc patients, and 91 MCInc patients. The Iris pipeline \cite{bron2021cross} was used to obtain modulated gray matter (GM) maps. In order to name patches with intuitive names, we named the image patches after the brain regions that highly overlapped with them based on the Hammers brain atlas \cite{hammers2003three}.\par

\subsection{Comparison Study}
We compared the performance of the \textit{GL-ICNN} against several baseline models, including black-box models such as \textit{GL-CNN}, VGG, and DenseNet. Additionally, to investigate the added value of the CNN feature extractors compared to simpler handcrafted features, we trained an EBM using brain GM volumes as input (\textit{Vol-EBM}). And we also investigated the added value of EBM compared to a simpler linear model that also would be interpretable as it allows to compute feature importance straightforwardly from the linear weights. To this end, we trained another glass-box model using the same CNN structure as the \textit{GL-CNN}, but replaced the output block with a Linear layer (\textit{GL-ICNN-L}). Furthermore, we compared the performance and computing time of the \textit{GL-ICNN} with the non-end-to-end {\textit{Glo\&Loc-EBM}}. We trained the models for five repetitions and recorded the average training time. 

\subsection{Validation of model performance} 
We validated the performance of the proposed model on AD diagnosis and MCI conversion prediction tasks, and on ADNI and PND cohorts. For the validation on ADNI cohort, subjects were randomly split into a training set, a validation set, and a testing set in a ratio of 8:1:1 in a stratified way according to class ratio. We trained the \textit{GL-ICNN} from scratch on AD-CN task and fine-tuned the model on MCIc-MCInc task. The PND cohort was used as external testing set.\par
To evaluate the performance of proposed models on binary classification tasks, we used Area Under the ROC Curve (AUC). 95\% Confidence Intervals (CIs) on the performance metrics were obtained based on 100 repetitions of bootstrapping on the testing set. And 95\% CIs of feature importance were obtained based on 100 repetitions of bootstrapping on the training set. We used DeLong’s test for determining whether the AUCs of two models were statistically significantly different.

\section{Results} 
\subsection{Model performance} 

\begin{figure*}[h]
\centering
\includegraphics[width=0.75\textwidth]{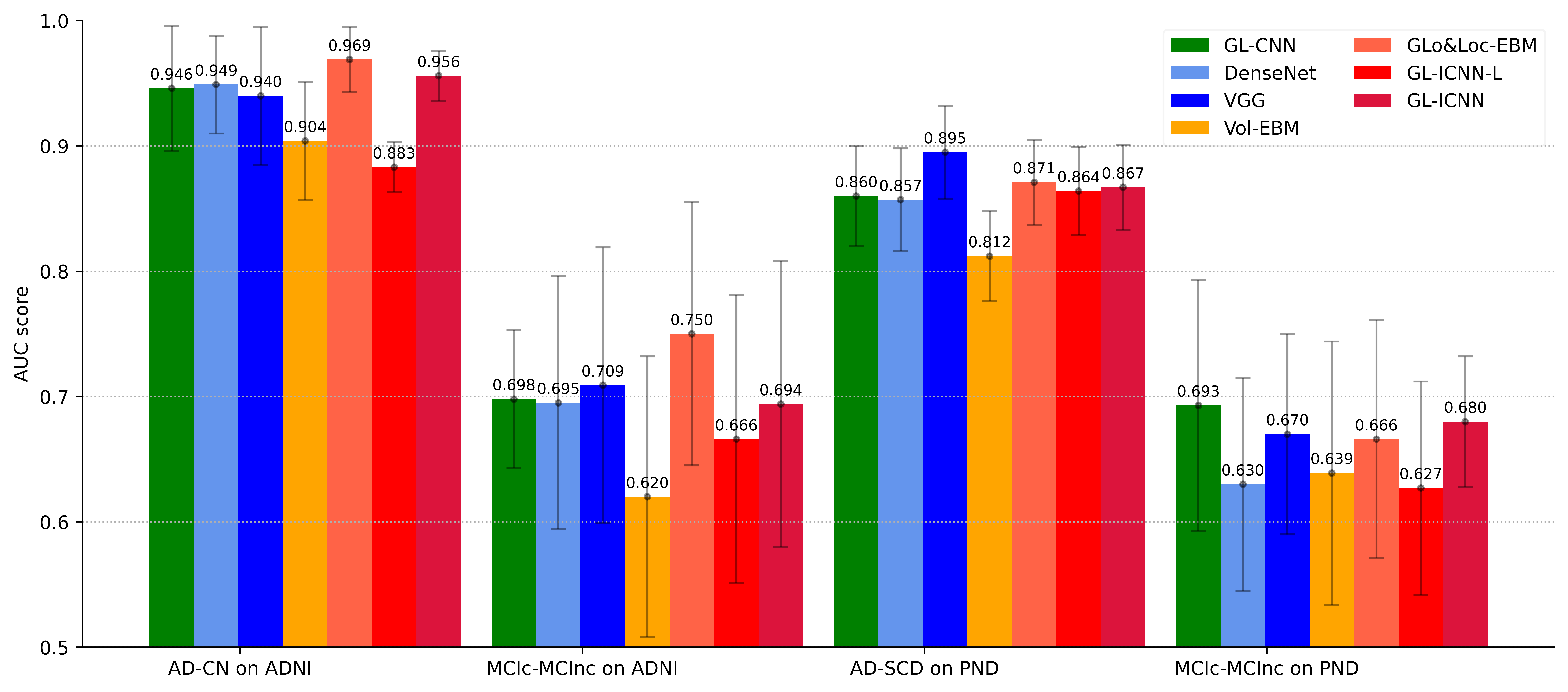}
\caption{The AUC score of models on 4 tasks. The first 3 models shown in cool colors are black-box models, while the last 3 models shown in warm colors are glass-box models, including the proposed \textit{GL-ICNN} displayed in red. The error bars represent the confidence intervals.} 
\label{fig performance}
\end{figure*}

The performance of the proposed and baseline models is shown in Fig.\,\ref{fig performance}.
The AUC of the \textit{GL-ICNN} is not significantly different from that of the black-box models and the \textit{Glo\&Loc-EBM} on any of the 4 tasks. For AD-CN task on ADNI, the AUC of \textit{GL-ICNN} (0.956) is significantly higher than the glass-box \textit{Vol-EBM} (AUC=0.904, p=0.04) and \textit{GL-ICNN-L} (AUC=0.883, p=0.01). For MCIc-MCInc task on ADNI, the AUC of \textit{GL-ICNN} (0.694) is not significantly different from that of the other models. The performance of the \textit{GL-ICNN} demonstrates a level of generalizability comparable to that observed in previous studies using the same cohorts \cite{kang2023interpretable,bron2021cross}. The total training time for \textit{Glo\&Loc-EBM} on the AD-CN task using the ADNI cohort was 81.3 hours, compared to 7.8 hours for \textit{GL-ICNN}. This significant reduction is because the end-to-end \textit{GL-ICNN} eliminates the need for a separate feature selection step and training multiple CNNs for feature extraction.

\subsection{Feature Importance for Interpretability}
\begin{figure}[h]
\centering
\includegraphics[width=0.45\textwidth]{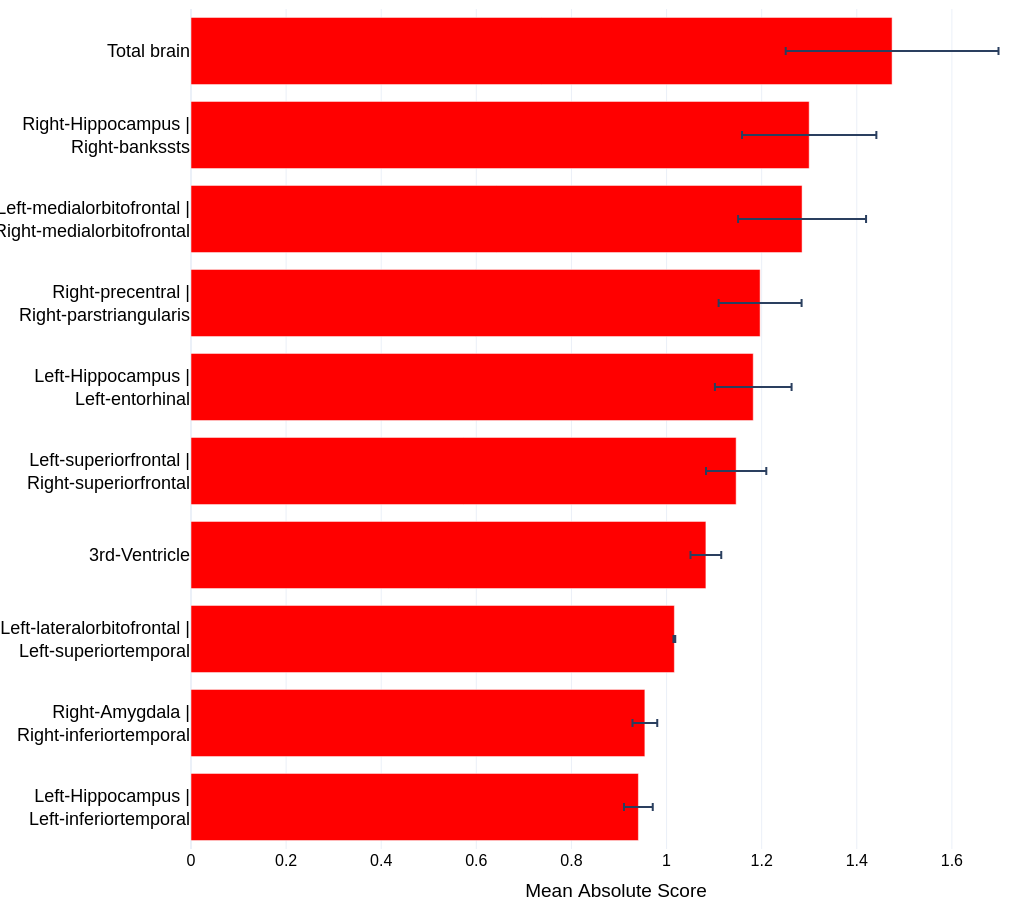}
\caption{The group-level feature importance of the \textit{GL-ICNN} for AD-CN task. The top 10 image patches of high importance are shown. The error bars represent for the confidence intervals of features.} 
\label{group}
\end{figure}

The group-level feature importance for the training set of the AD-CN task in the ADNI cohort is shown in Fig.\,\ref{group}. We observed that the \textit{GL-ICNN} based the diagnosis mostly on regions of the temporal lobes, especially all regions containing the hippocampus and amygdala appeared for AD-CN task on ADNI. The brain regions with high feature importance align with prior clinical findings on AD \cite{jellinger2020towards}. We plan to collaborate with clinical experts to validate the reliability of the feature importance identified by the \textit{GL-ICNN}.

\par

\section{Conclusion}
The main contribution of this work is the introduction of a novel model architecture and training strategy that enables the model to directly take imaging data as input and learn the imaging features in an end-to-end fashion while remaining interpretable. The \textit{GL-ICNN} achieves a comparable performance with state-of-the-art black-box models. In addition, the \textit{GL-ICNN} provides the feature importance of brain regions to show how the model makes decisions.



\bibliographystyle{IEEEtran}
\bibliography{reference}


\end{document}


\setlength{\textfloatsep}{5pt} 

\section*{Supplementary Materials}

\subsection{CNN structure}

\begin{figure*}[h]
\centering
\includegraphics[width=0.9\textwidth]{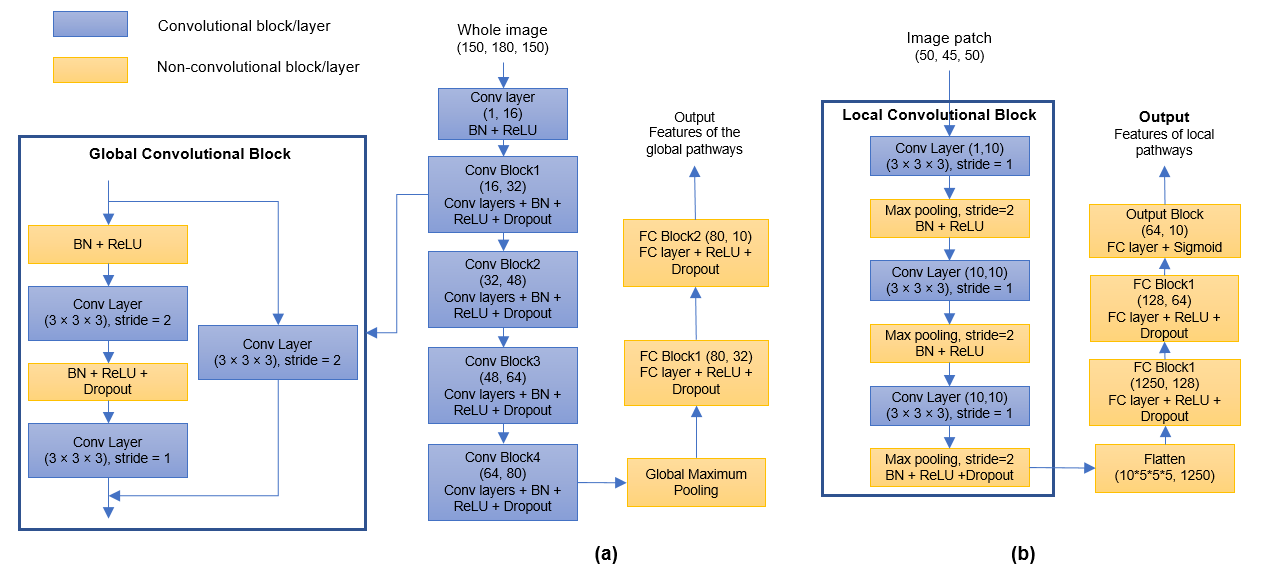}
\caption{The CNN structure of the (a) \textit{Glo-CNN} backbone (b) \textit{Loc-CNN} backbone.} 
\label{cnn}
\end{figure*}

The CNN structure of the global CNN backbone which takes whole image as input, and the local CNN backbone backbones which take image patches as input, are shown in Fig\,\ref{cnn}. The global CNN backbone is a DenseNet \cite{huang2017densely}, and the Local CNN backbone is a VGG \cite{simonyan2014very}.

\subsection{Definition of Feature Importance}
The pseudo-code for training EBM is shown in the Algorithm\,2 of \cite{nori2021accuracy}. Because of the ability of EBM to analyze features independently, it is also straightforward to interpret the importance of each feature to the prediction \citep{caruana2015intelligible}. Hence, the individual-level feature importance vector for subject $i$ is defined as the collection of its shape function outputs:

\begin{equation}\label{eqn-2}
I_{indiv}(X_i) = \left(f_1(x_1), \ldots, f_k(x_k) \right)  
\end{equation}
\noindent
and group-level feature importance vector for a dataset $D$ is defined as the average of the absolute individual-level feature importance measures:
\begin{equation}\label{eqn-3}
I_{group}(D) = \frac{1}{N} \sum_{i=1}^{N} |I(X_i)|  
\end{equation}
\noindent
where $|\cdot|$ denotes the element-wise absolute operator.
\par

\subsection{Study materials}

We included T1-weighted (T1w) MRI scans from the Alzheimer’s Disease Neuroimaging Initiative (ADNI) and the Health-RI Parelsnoer Neurodegenerative Diseases Biobank (PND) cohorts. The ADNI cohort consists of 335 AD patients, 520 control participants (CN), 231 mild cognitive impairment (MCI) patients who converted to AD within 3 years (MCIc), and 629 MCI patients who did not convert (MCInc) within 3 years. The PND cohort is a collaborative biobanking initiative of the eight university medical centers in the Netherlands \citep{mannien2017parelsnoer}, we included 198 AD patients, 138 participants with SCD, and 139 MCI patients. Of the MCI group, we included the participants that had a follow-up period of at least 6 months. 48 MCI patients converted towards AD within the available follow-up time and 91 MCI patients remained stable. The PND cohort differs from the ADNI cohort in population distribution, patient inclusion criteria, and diagnostic criteria \cite{aalten2014dutch}.
\par

The Iris pipeline \cite{bron2021cross} was used to obtain modulated gray matter (GM) maps. First, a group template space was created using pairwise registration. Then, probabilistic GM maps were generated using SPM8 \citep{ashburner2005unified}, transformed to the group
template space, modulated for compression and expansion, and normalized by intracranial volume. The volumetric measures for ADNI and PND cohorts are also collected.
\par

In order to name patches with intuitive names, we calculated the overlaps between the image patches and brain regions on the Hammers brain atlas \cite{hammers2003three}, and named the image patches after the brain regions that highly overlapped with them.\par

\subsection{Feature Importance for Interpretability}

\begin{figure*}[h]
	\centering  
	\subfigbottomskip=2pt 
	\subfigcapskip=-5pt
  	\subfigure[Group-level feature importance for MCIc-MCInc task on ADNI]{
		\includegraphics[width=0.48\linewidth]{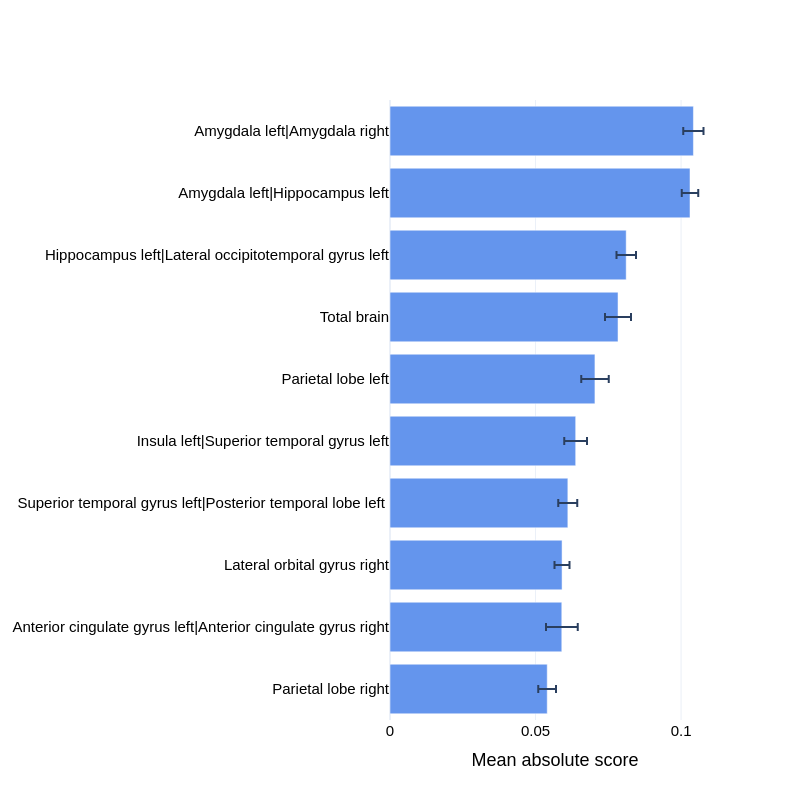}}
        \subfigure[Group-level feature importance for AD-SCD task on PND]{
		\includegraphics[width=0.48\linewidth]{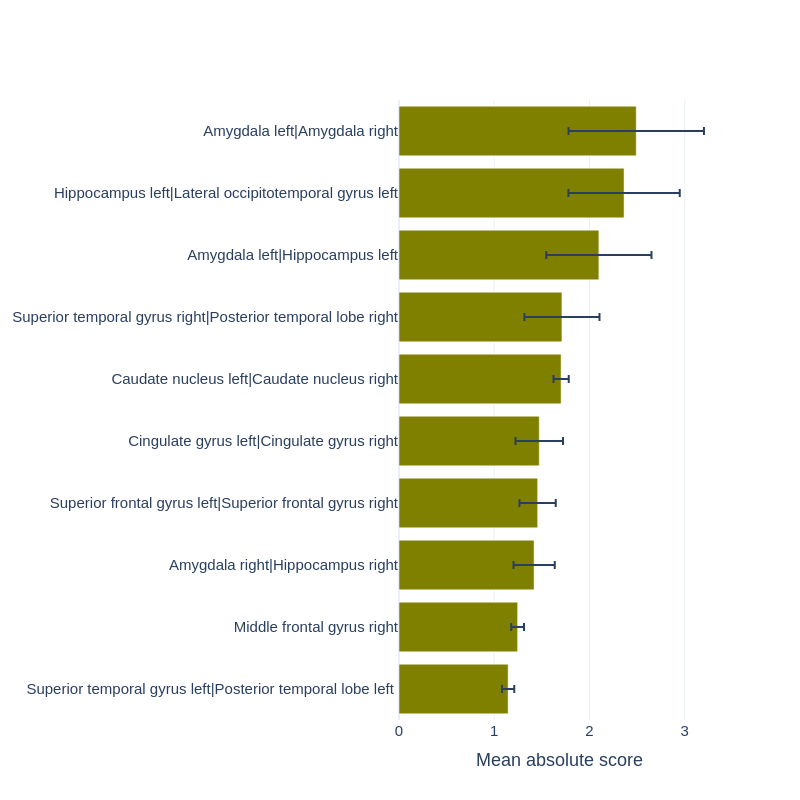}}
	\subfigure[Individual-level feature importance]{
		\includegraphics[width=0.48\linewidth]{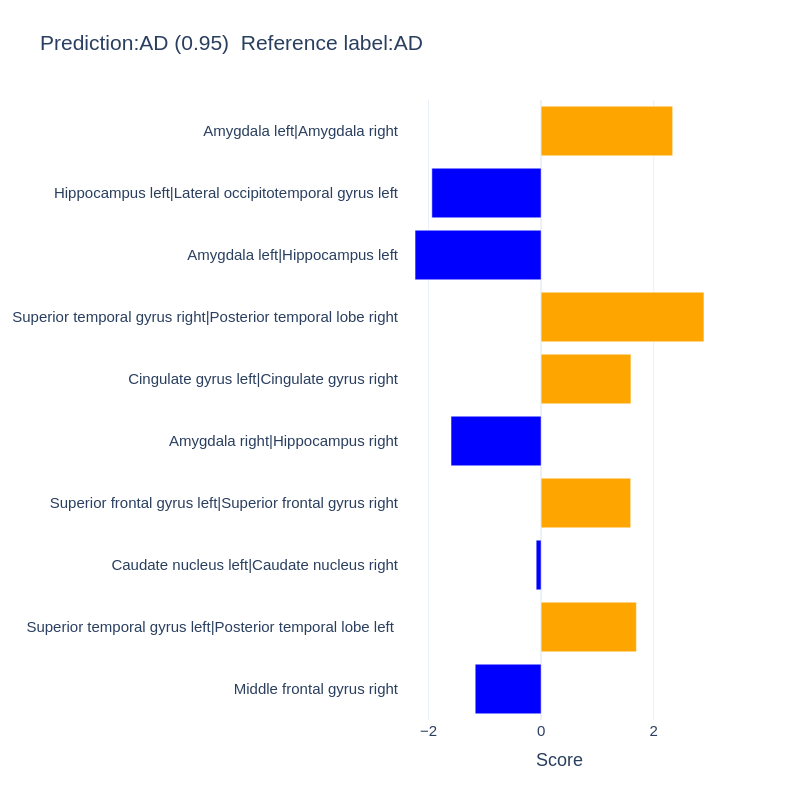}}
	  \\
	\caption{The group-level feature importance of the \textit{GL-ICNN} for (a) MCIc-MCInc task on ADNI, and (c) AD-SCD task on external testing set PND. (c) The individual-level feature importance of a correctly classified AD patient on ADNI. The TOP 10 image patches of high importance are shown. The error bars represent for the confidence intervals of features in the group-level feature importance. For the individual-level feature importance, the predicted label, posterior probability and reference label are shown. The positive number in orange shows the feature contributes to the positive diagnosis, and the negative number in blue shows the feature contributes to the negative diagnosis in the individual-level feature importance.}
\label{importance}
\end{figure*}

The group-level feature importance is shown in Fig.\,\ref{importance}(a-b). The distribution of TOP 10 image patches is more dispersed on MCIc-MCInc compared to AD-CN task, while hippocampus and amygdala are still included in Fig.\,\ref{importance}(a). For the group-level feature importance on AD-SCD task on external PND testing set, we reached similar conclusions as the AD-CN task on ADNI that the regions of the temporal lobes still have high feature importance (Fig.\,\ref{importance}(b)). The individual-level feature importance is shown in Fig.\,\ref{importance}(c), it shows how model made decisions on each data sample.

\bibliographystyle{IEEEtran}
\bibliography{reference}
\